\pdfoutput=1
\documentclass[11pt]{article}

\usepackage{booktabs} % For professional looking tables
\usepackage{multirow} % For multirow cells

\usepackage[final]{acl}

\usepackage{times}
\usepackage{latexsym}

\usepackage{dsfont}
\usepackage{amsmath}

\usepackage[T1]{fontenc}
\usepackage[utf8]{inputenc}

\usepackage{microtype}

\usepackage{inconsolata}
\usepackage{graphicx}

\title{SAKE: Steering Activations for Knowledge Editing}

\author{
 \textbf{Marco Scialanga\textsuperscript{1,*,$\dagger$}},
 \textbf{Thibault Laugel\textsuperscript{2,3,*}},
 \textbf{Vincent Grari\textsuperscript{2,3}},
 \textbf{Marcin Detyniecki\textsuperscript{2,3,4}}
\\
 \textsuperscript{1}EPFL, Lausanne, Switzerland\\
 \textsuperscript{2}AXA, Paris, France,\\ 
 \textsuperscript{3}TRAIL, LIP6, Sorbonne Université, Paris, France\\
 \textsuperscript{4} Polish Academy of Science, IBS PAN, Warsaw, Poland
\\
 \small{
    \textsuperscript{*}Equal contribution ;
    $^\dagger$Work conducted during an internship at AXA} \\
   \small{\textbf{Correspondence:} \href{mailto:marco.scialanga@epfl.ch}{marco.scialanga@epfl.ch}, 
   \href{mailto:thibault.laugel@axa.com}{thibault.laugel@axa.com}
   }
 }

\begin{document}

\maketitle

%%%%%% Abstract %%%%%%
\begin{abstract}
As Large Langue Models have been shown to memorize real-world facts, the need to update this knowledge in a controlled and efficient manner arises. Designed with these constraints in mind, Knowledge Editing (KE) approaches propose to alter specific facts in pretrained models. However, they have been shown to suffer from several limitations, including their lack of contextual robustness and their failure to generalize to logical implications related to the fact. To overcome these issues, we propose SAKE, a steering activation method that models a fact to be edited as a distribution rather than a single prompt. Leveraging Optimal Transport, SAKE alters the LLM behavior over a whole fact-related distribution, defined as paraphrases and logical implications. Several numerical experiments demonstrate the effectiveness of this method: SAKE is thus able to perform more robust edits than its existing counterparts. The code to reproduce all experiments is made available on a repository\footnote{\url{https://github.com/axa-rev-research/knowledge-editing}}.
\end{abstract}

%%%%%% Main Text %%%%%%
%A few notes on the main text:  
%Three heading levels are permitted. Only the headings listed are permitted as Level 1 headings. Authors are encouraged to indicate the level of each heading using a unique format. For example, 
%\textbf{LEVEL 1 IN BOLD CAPS}
%\textbf{Level 2 in bold}
%\textit{Level 3 in italics}

\section{Introduction}
It is well known that Large Language Models (LLMs) can store numerous facts in their parameters and recall them when prompted accordingly~\cite{petroni2019language,jiang2020can,chang2024llm_factual_knowledge,DBLP:conf/naacl/WangHBP24}. 
%However, the training process needed to make LLMs learn these facts is extremely expensive in terms of computational resources, due to both the high number of parameters and the large amount of text needed to train high-performing models~\cite{DBLP:conf/nips/BrownMRSKDNSSAA20}. 
%On the other hand, due to their ability to generate text on virtually every topic (though not without pitfalls), LLMs are being widely used as, for example, personal assistants or chatbots. 
As they are generally used in a dynamic environment, the need of keeping these models up to date with new information and erasing obsolete facts from their memory emerges. 
Traditionally, this has been accomplished with parameter fine-tuning~\cite{zhu2020modifying}, optimizing the pre-trained weights of LLMs with new data. %Even though efficient and constrained fine-tuning techniques exist~\cite{hu2021lora,rafailov2024direct, DBLP:conf/emnlp/HuWLXLB0PL23}, these can still be resource intensive and induce overfitting~\cite{DBLP:conf/nips/YangDQWN22}. 
Yet, despite efforts to make them more efficient~\cite{hu2021lora,rafailov2024direct},  the high computational cost and significant risk of overfitting make fine-tuning techniques being generally viewed as being suboptimal for the task of facts editing~\cite{meng2022locating}.%, which requires a more efficient and controlled approach.

Consequently, several recent works have focused on Knowledge Editing (KE), which aims to precisely alter specific facts in the memory of LLMs through a list of edits (see~\citet{wang2023knowledge} for a survey). %strong performance across the main key desiderata~\cite{meng2022locating}.
%: accuracy, which refers to the precise recall of the edit; generality, which ensures the edit is successfully applied to its paraphrased forms; and specificity, which prevents unintended alterations to unrelated prompts. 
However, these methods have been shown to suffer from several limitations: (1) One well-known challenge is that edited LLMs often struggle to generalize their newly acquired knowledge to various types of logical implications \cite{cohen2024evaluating}. (2) Edited LLMs are often not robust to certain rephrasing of the prompts, including those that could realistically arise in a conversation, such as long, noisy prompts or prompts raising doubts on the edited knowledge \cite{ma2024robustness}. 
(3) Finally, they generally lack flexibility in revising or removing prior edits. Approaches that modify model weights, such as ROME~\cite{meng2022locating} and MEMIT~\cite{meng2022mass}, or rely on external memory in the form of auxiliary networks~\cite{mitchell2021fast,mitchell2022memory}, do not inherently support efficient revision or removal of prior edits, thus making continuous updates challenging.

\begin{figure}
    \centering
    \includegraphics[width=0.9\linewidth]{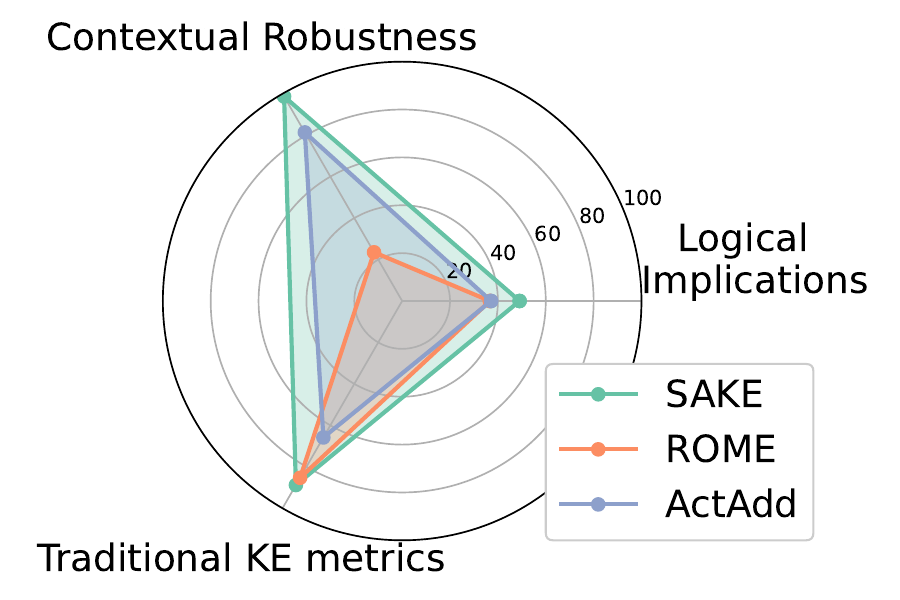}
    \caption{Summary of the empirical results obtained with SAKE (our method) compared to a weight editing method (ROME) and an activation steering method (ActAdd) for GPT2-XL.}
    \label{fig:spider-chart}
\end{figure}

In this paper, we argue that these limitations come from the design of most KE methods. 
Most approaches indeed aim to alter the behavior based on a single input prompt, therefore leading to overfitting and poor generalization properties. 
After  providing background and discussing this issue in Section~\ref{sec:background}, we propose, to address it, a novel fact-editing framework, called \textbf{S}teering \textbf{A}ctivations for \textbf{K}nowledge \textbf{E}diting (SAKE). Described in Section~\ref{sec:proposition}, SAKE is based on activation steering and is intended to provide greater flexibility, robustness, and control over knowledge modifications. Unlike traditional KE methods, our approach optimizes the LLM behavior over distributions of text, rather than individual prompts, allowing for more generalizable and consistent edits. 
%Furthermore, it ... ensuring that updates remain targeted and minimally invasive to unrelated knowledge. 
Our experimental results, described in Sec.~\ref{sec:experiments} and summarized in Figure~\ref{fig:spider-chart}, 
demonstrate that our proposed method outperforms existing KE approaches in terms of generalization to logical implications, contextual robustness, as well as traditional KE metrics, hence showcasing its capability to perform more robust edits overall.

% \textbf{Examples?}
% \begin{itemize}
%     \item LLM are seen as some kind of world-knowledge database.
%     \item Hence need to update facts, when not up-to-date, etc.
%     \item Most existing methods do blabla
%     \item Yet, these theories fail in practice, due to various reasons:
%     \begin{itemize}
%         \item to do.
%         \item for instance, the fact that we frame the prompt and not knowldege, and therefore paraphrases etc. fail
%     \end{itemize}
%     \item To illustrate this problem, 
%     \begin{itemize}
%         \item we rephrase a sentence and it does not work with ROME anymore? Subject Alias? Comp?
%     \end{itemize}
%     \item Noted by various works in the literature: works on robustness, ripple effects. 
%     \item On the other hand: external memory but too heavy. And the difference steering.
%     \item We therefore propose a new method to edit facts through activation steering has the following benefits
%     \begin{itemize}
%         \item take into account (easily controllable) distributions, instead of just a prompt, allowing robust edits
%         \item scopes edit in a specific phrase, leaving the network overall unchanged, and....?
%     \end{itemize}
% \end{itemize}

\section{Background}
\label{sec:background}

\subsection{Knowledge Editing}
%The popularity of LLMs has surged in recent years, with users relying on the knowledge stored in their parameters when interacting with personal assistants or chatbots. Thus, it is necessary to find efficient and reliable ways to keep LLMs updated, making them forget obsolete knowledge and learn new, true facts. While fine tuning techniques could be used to edit , they are often still computationally expensive might induce catastrophic forgetting of relevant knowledge and / or overfitting 
Knowledge Editing (see~\citet{wang2023knowledge} for a survey) is the task of modifying the knowledge stored in LLMs.
Given a language model $f$, an \textbf{edit} is generally defined by a source tuple $(s,r, o)$ ($s$ being the \emph{subject}, $r$ the \emph{relation}, and $o$ an old \emph{object}), and a target tuple $(s, r , o^*)$, with $o^*$ a new object. Concretely, the object $o$ is to be returned when querying the model with $(s,r)$ in the form of natural language (the prompt $p$): 
the goal of model editing is thus to define an edited model $f^*$ that outputs $o^*$ given the prompt $p$, an objective generally defined as "edit success" or "accuracy". 
%Given a generator model $G$, an \textbf{edit} is made of a prompt $p = (s,r)$ ($s$ being the \emph{subject}, $r$ the \emph{relation}), an old \emph{object} $o$ (we assume $o=\arg\max_x G(x|p)$), and a new object $o^*$. 
Besides accuracy, two other objectives are generally associated to the task of knowledge editing~\cite{meng2022locating,wang2023knowledge}: the ability 
of the model to generate $o^*$ given paraphrases $p'$ of $p$ (generality), and how well the initial behavior of $f$ is preserved for unrelated prompts (specificity).
%Traditionally, the three primary goals of KE are to obtain, after the edit $\arg\max_x G^*(x|p) = \arg\max_x G^*(x|p') = o^*$ where $G^*$ is the edited model and $p'$ is any paraphrase of $p$ (accuracy, generality), and to not deviate from the behavior of $G$ for any unrelated prompt (specificity).
%\paragraph{probably here: "traditional" methods for KE}

\paragraph{Traditional methods for KE} Existing knowledge editing (KE) techniques employ a range of strategies. Methods based on targeted modifications of the model's weights, such as ROME~\cite{meng2022locating} and MEMIT~\cite{meng2022mass} offer the advantage of preserving the original inference procedure. However, the knowledge localization assumption they rely on has been the topic of criticism~\cite{hase2024does}, as well as their tendency to overfit the input prompt~\cite{zhang2024uncovering}.
%may exhibit unpredictable behaviors with specificity prompts.
%ROME and MEMIT also have the need of specifying a subject, often incurring in the risk of overfitting on that specific token. 
Instead of targetting a specific localization, the change in parameters for a given edit are predicted using a hypernetwork in the case of MEND~\cite{mitchell2021fast}.

Other methods rely on external memory units. This is the case of GRACE~\cite{hartvigsen2024aging}, which uses a memorized codebook to map relevant activations to values that increase the likelihood of the desired output.
Similarly, SERAC~ \cite{mitchell2022memory}, passes an input that is to be edited to an external fine-tuned model. 
%both an in-scope classifier to determine when an input is in the scope of the edit, in which case it is passed to an external fine-tuned model.
Unfortunately, using additional models for memory (and edit scope detection in the case of SERAC) is expected to bring additional training overhead, as well as longer inference time.

Finally, in-context strategies such as IKE \cite{zheng2023can}, modify the model’s behavior temporarily during inference by adding additional context in the form of instructions and examples.
Although performing well across traditional KE benchmarks, in-context strategies have been shown to struggle in more realistic scenarios~\cite{ma2024robustness}, and are generally expected to suffer from in-context learning (ICL) limitations in terms of generalization~\cite{mosbach2023few}.
%long noisy prompts, contradictions, or misleading information.

% Still missing: general comment about how good they are in general, and/or how they compare. To discuss. 

% \begin{table*}
% \centering
% \resizebox{\linewidth}{!}{%
% \begin{tabular}{c|c|c}
%     Category & Identified issues & Papers \\
%     \midrule
%     Logical Implications & Compositionality I, Compositionality II, Subject Aliasing, Relationship Specificity, Logical Genearlization/Symmetry & \cite{ma2023untying}\\
%     Contextual Robustness & Raising doubt, long context, dialogue & \\ 
% \end{tabular}}
%     \caption{Caption}
%     \label{tab:my_label}
% \end{table*}

\subsection{Three Limitations of Existing Methods}
\label{sec:limitations}

%Although existing methods can achieve satisfying results over the aforementioned properties, numerous works have pointed out issues with the traditional formalization of the editing task. 
In this section, we identify three main limitations, common to all aforementioned KE methods, that we seek to address in this paper.

\paragraph{Logical implications}
One implicit, but crucial, objective of knowledge editing is that altering a fact in a model should affect all knowledge directly connected to this fact as well. As a result, it is expected that answers to questions on logical implications deriving from the edit should reflect this change as well. These could include, given e.g. an edit performed to update the name of the US president, the composition of multiple relations (e.g. "the son of the US president") or subject aliasing ("the head of state in the US").
%, and the generalization of the edit to long, noisy prompts. 
%A common example is by updating the name of the president of the USA, the name of the "son of the president of the USA" should change as well. 
Yet, several works~\cite{ma2023untying,cohen2024evaluating,liunveiling} have pointed how edited models with existing approaches often fail to generalize their updates to their respective logical implications. 
In particular, following the categorization of logical implications proposed by~\citet{cohen2024evaluating}, edited models have been shown to struggle with prompts arising from the composition of two relations (\emph{Compositionality I} and \emph{II}), and to a lesser extent even from just using aliases of the subject (\emph{Subject Aliasing}). 
Moreover, the authors have found that edited models often struggle with prompts with the same subject but a different relation than those in the list of edits, deviating from the original output when it is actually not requested (\emph{Relation Specificity}). 

\paragraph{Contextual robustness} 
Besides, the edits performed with existing methods have been shown to lack contextual robustness. \citet{ma2024robustness} thus observe how, in a realistic conversational setting, it is quite easy to make the edited model doubt its newly acquired knowledge, and even to revert it to the pre-edit behavior. Additionally, they have found that edited model struggle with long or noisy contexts, and in situations with prompts raising doubts. Similarly to logical implications, this lack of contextual robustness is problematic as knowledge edits are generally intended to hold across diverse situations, e.g. to patch the behavior of a customer-facing chatbot with updated information.

% \paragraph{here to do: why it fails. where does this shortcoming come from. Preparing the way for "why our method will help".}

% \begin{itemize}
%     \item Logical Implications + Contextual Robustness can be viewed as the subspace we would like our edit to generalize to. 
%     \item Notations, we write the 
%     \item As pointed out, existing methods focus on the fact as a prompt $p$, and do not perform this change robustly
%     \item 
% \end{itemize}

\paragraph{Flexible editing mechanism}
On a different note, we argue that another limitation of existing methods is their lack of flexibility in adding, removing, or altering edits. As knowledge needs to be continuously updated, the ability to update the information stored in a flexible way is crucial. Yet, as multiple edits are being performed simultaneously, methods relying on direct intervention on the model weights such as ROME~\cite{meng2022locating} and MEMIT~\cite{meng2022mass}, and approaches relying on the training of an external memory network such as MEND~\cite{mitchell2021fast} and SERAC~\cite{mitchell2022memory} are unable to simply undo a specific edit. 
Worse, performing the inverse edit $(s,r,o^*\to o)$ to restore the previously lost knowledge has been shown not only to fail in bringing the model back to its original form, but also severely hurt the model's overall performance~\cite{liunveiling,hu2024wilke}.
Since one of the main advantages of knowledge editing over traditional model fine-tuning is the ability to preserve the original model, its limitations in this regard present a significant challenge.

\section{Proposition}
\label{sec:proposition}

To overcome the three limitations discussed in the previous section, we introduce in this section a new knowledge editing framework.
After formalizing a new objective in Sec.~\ref{sec:formalization-OT}, we introduce in Sec.~\ref{sec:sake} \textbf{S}teering \textbf{A}ctivations for \textbf{K}nowledge \textbf{E}diting (SAKE), our approach designed to tackle the issues identified.
%in Sec.~\ref{sec:limitations}. SAKE relies on modifying input activations at test time for each edit independently, thus allowing for a very flexible editing mechanism.

%With SAKE, we cover the in-scope input space with a \textit{source distribution} of activations that are associated with (and not limited to) $(s,r,o)$ and map it to a \textit{target distribution} made of activations arising from the consequences of $e = (s,r,o \to o^*)$. At inference, we map activations from the source to the target distribution, using optimal linear transport, when these fall into the in-scope input space.

\subsection{Knowledge Editing as a Distribution Mapping Problem}
\label{sec:formalization-OT}
One interpretation for the inability of existing KE methods to generalize well to related inputs (other than simple paraphrases) is that the knowledge to edit is generally represented (and thus passed in the model) as an isolated \textit{prompt}. This results in a discrepancy between how humans perceive a fact, composed of high-level concepts associated the subject $s$, the relation $r$, and the object $o$, and what is actually passed to the model (a prompt). As a result, the edit may fail to generalize to mere reformulations of the sentence, let alone some logical implications.
To address this issue, we propose to define the scope of an edit $e$ to include these notions. Following the notations from~\citet{wang2023knowledge}, we thus introduce~$\mathcal{X}_e  \subset \mathcal{X}$ as the in-scope input space, $\mathcal{Y}_e \subset \mathcal{Y}$ as the original output space and $\mathcal{Y^*}_e \subset \mathcal{Y}$ as the target output space. Concretely, $\mathcal{X}_e$ covers all the inputs which output should be affected by the edit $e$. As $e$ is a piece of factual knowledge, these include paraphrases of the prompt, but also any logical implication, or contextual knowledge.
The objective of a single edit $e$ is then defined by the following robust optimization problem: 
\begin{equation}
    \begin{aligned}
        \underset{f^*\in\mathcal{F}}{\min} \mathds{E}_{x, y^* \in \mathcal{X}_e, \mathcal{Y^*}_e} \mathcal{L}(f^*(x), y^*) \\
       \text{s.t. }f^*(x) = f(x) \, \forall \,x \in \mathcal{X} \setminus \mathcal{X}_e,
    \end{aligned}
    \label{eq:robust-editing-objective}
\end{equation}
where $f^*$ is the edited model, $y^*$ is the new desired output (note: not necessarily $o^*$), $\mathcal{L}$ a loss assessing the success of the edit by measuring the difference between the edited model’s and the desired output; %$y^*$
and $\mathcal{F}$ is the set of all possible edited models.
Considering facts are interconnected by a plethora of possible logical implications, it is clear that $\mathcal{X}_e$ is quite large, especially when compared to the mere prompt $(s,r)$. This underlines the importance to account for the whole distributions of $\mathcal{X}_e$ and $\mathcal{Y}_e$ to perform more robust edits.
%$\mathcal{X}_e$ might contain, for example any combination of different subject, relation, old or new object when compared to $e = (s,r,o\to o^*)$. 
%any method that solely relies on the prompt $(s,r,o\to o^*)$ to perform the edit is destined to fail at sufficiently covering $\mathcal{X}_e$, and so at generalizing to logical implications and different types of prompts other than just the plain $(s,r)$ and its clean paraphrases. 

To approximate the set~$\mathcal{X}_e$, we aim to generate a set of inputs that are affected by the edit $e$ (i.e. not just $(s,r,o)$ or its paraphrases). It is clear now that our method cannot simply be based on mapping $o \to o^*$, because the completions of the inputs in $\mathcal{X}_e$ are not limited to the original objects of the edit.

Thus, we need to find a mapping from a distribution of obsolete knowledge connected to the fact $(s,r,o)$, which is exactly $\mathcal{Y}_e$, to a new one revolving around $(s,r,o^*)$, essentially $\mathcal{Y^*}_e$. This mapping should be applied whenever a new input at inference time belongs to $\mathcal{X}_e$. As a result, we define our edited model to be $f^* := W(m(T(x)))$, where $W$ is the linear modeling head of $f = W\circ T$ ($T$ standing for transformer) and $m$ is a function that maps the activations in the last layer $h^s$ from a source distribution $\mathcal{S}_e$ (describing the obsolete knowledge) to $h^t$ in the target distribution $\mathcal{T}_e$ (updated knowledge) . 

With this new viewpoint, equation \ref{eq:robust-editing-objective} thus can be reformulated as:
\begin{equation}
    \begin{aligned}
        \underset{m: \mathbf{R}^d\to\mathbf{R}^d}{\min} \mathds{E}_{h^s, h^t \in \mathcal{S}_e,\mathcal{T}_e} \mathcal{L'}(m(h^s), h^t) \\
       \text{s.t. }m(h) = h \ \forall \ h \in \mathbf{R}^d \setminus \mathcal{S}_e,
    \end{aligned}
    \label{eq:robust-editing-objective-activations}
\end{equation}
where $\mathcal{L'}$ is an appropriate loss function comparing the transformed layer to the target layer and $d$ the dimensionality of the activation space of $f$. In the next subsections we accurately describe how we collect the activations and how we learn the optimal mapping $m$, leading to the complete formulation of the knowledge editing method we propose.

% TODO THibault: Considering facts are interconnected, it is clear that $\mathcal{X}_e$ is quite large, especially when compared to the mere prompt $(s,r)$. 
% $\mathcal{X}_e$ might contain, for example any combination of different subject, relation, old or new object when compared to $e = (s,r,o\to o^*)$ (we could put here an example with different s,r, or o when we change the previous one). It is clear then, that any method that solely relies on $(s,r,o\to o^*)$ to perform the edit is destined to fail at sufficiently covering $\mathcal{X}_e$, and so at generalizing to logical implications and different types of prompts other than just the plain $(s,r)$ and its clean paraphrases. 

% \begin{itemize}
%     \item From there we write $f'(x)$ as $h'(g(x))$
%     \item Thus h' is a mapping between distribution $g(X)$ and $Y$
% \end{itemize}

\subsection{SAKE: Steering Activations for Knowledge Editing}
\label{sec:sake}

\begin{figure}
    \centering
    \includegraphics[width=1.0\linewidth]{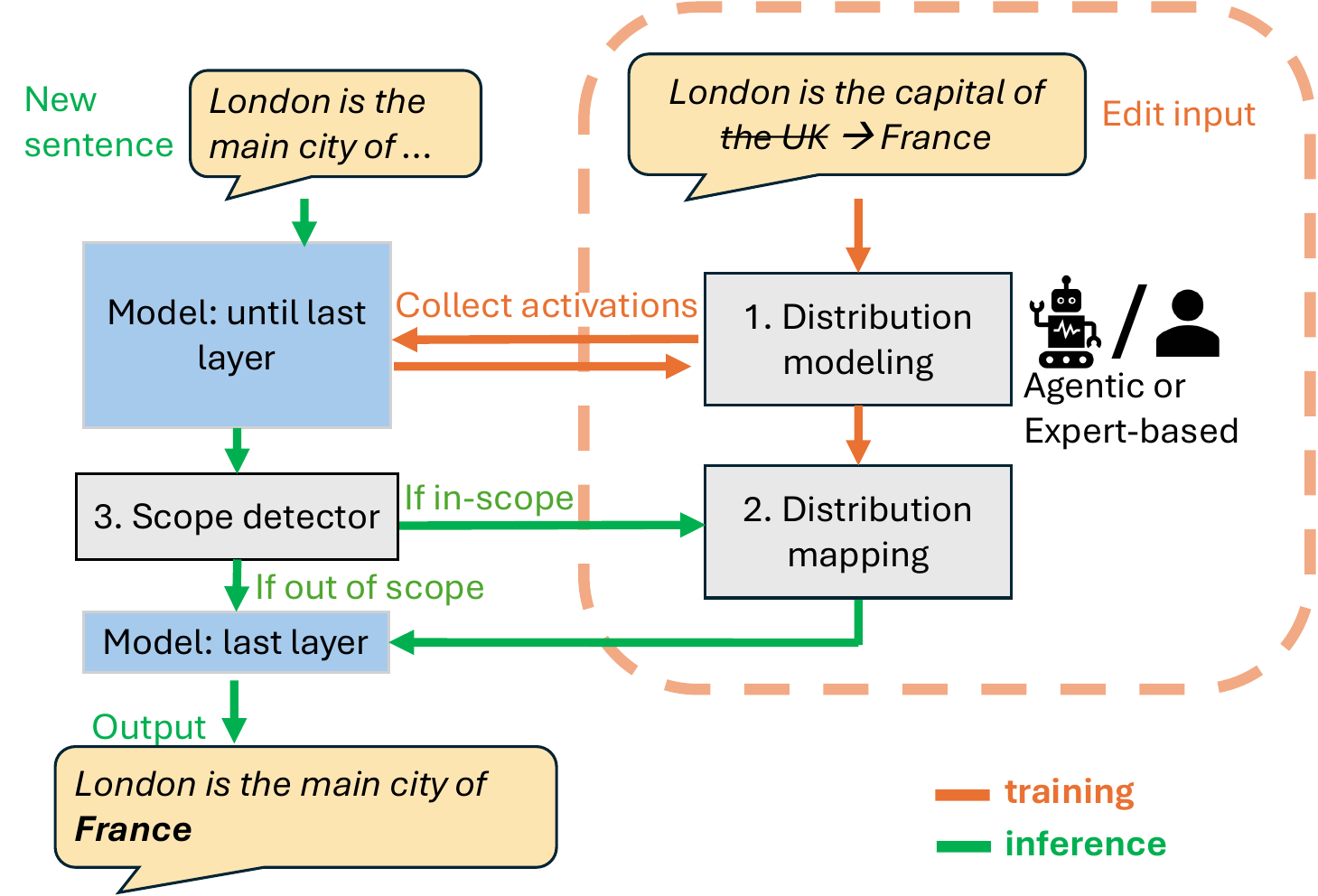}
    \caption{Overview of SAKE, illustrating our method's behavior both at training (orange) and test (green) times.
    }
    \label{fig:SAKE-algo}
\end{figure}

To solve the problem described in Equation~\ref{eq:robust-editing-objective-activations}, we introduce SAKE, which relies on activation steering to perform robust knowledge editing. SAKE relies on three components, which can be visualized as grey blocks in Figure~\ref{fig:SAKE-algo}, and each described in the following subsections. First, we propose to directly model $\mathcal{Y}_e$ and~$\mathcal{Y^*}_e$ in the activation space (Sec.~\ref{sec:sake-sampling}, number (1) in Fig.~\ref{fig:SAKE-algo}); then, we map these distributions using optimal transport (Sec.~\ref{sec:sake-mapping}, number (2)). For inference, we implement a scope detection mechanism to target a potentially relevant edit when an input $x \in \mathcal{X}_e$ is passed to the model (Sec.~\ref{sec:sake-scope}, number (3)).

\subsubsection{Modeling Source and Target Distributions}
\label{sec:sake-sampling}

% In order to perform activation steering, we first need to generate the in-scope source and target distributions. However, directly estimating these distributions in the activation space is difficult, \textbf{I mean how would one even do it? I can't think of another way, even if more difficult} as ARGUMENT TO DO????? REF.
% Therefore, we rely on modeling~$\mathcal{X}_e$ and~$\mathcal{Y}_e$ in the input space. Concretely, 
To model $\mathcal{X}_e$, as well as, indirectly, $\mathcal{Y}_e$ and $\mathcal{Y^*}_e$, we need sentences in natural language $P_e := \{p_i^e\}_{i\leq n}$ that are representative of the in-scope inputs that the edit $e$ should generalize to. This thus allows us to include paraphrases and logical implications in these distributions, but also can be adapted differently to fit specific contexts, such as question answering, text completion, or classification for instance.
We identify two strategies for modeling~$\mathcal{X}_e$, detailed below.

\paragraph{Agentic generation:} A natural way to generate these distributions is to rely on a trained LLM such as GPT-4, prompted with instructions to paraphrase and derive logical implications of the base edit. 
Previous works have shown the efficiency of generating paraphrase and specificity prompts for knowledge editing~\cite{zhang2024uncovering,gangadhar2024model}. In this work, we extend this to logical implication prompts. 
Examples of instructions are given in Appendix~\ref{appendix-input-prompts}.

\paragraph{Expert-based generation:} As the distributions are in the input space, the sentences can be directly generated by human users. In particular when the number of edits is low, 
it thus allows domain experts to directly define specify the fine-grained behaviors that are expected by the model after editing. This is especially interesting in applications where naive text generation would be insufficient~\cite{mayring2025qualitative}.

%To perform an edit with activation steering, we first need to generate source and target distributions. 
Since these sentences will also be used to cover $\mathcal{Y}_e$ and $\mathcal{Y^*}_e$ with $\mathcal{S}_e$ and $\mathcal{T}_e$ respectively, $p_i$ should be meant to be completed immediately with either $y \in \mathcal{Y}_e$ or $y^* \in \mathcal{Y^*}_e$, e.g. "The US president is", or "The son of the US president is". To collect the \textit{source activations} $\mathcal{S}_e$, we pass $p_i$ through $f$ and save the activation at the layer and token of choice. To collect the \textit{target activations} $\mathcal{T}_e$, we need to first build $\tilde{p}_i = (c, p_i)$, where $c$ is some context that ensures that the output of the unedited model when prompted with $\tilde{p}_i$ would belong to $\mathcal{Y^*}_e$ and not $\mathcal{Y}_e$. An example of $\tilde{p}_i$ for paraphrases could be: \textit{Do not mention} $o$\textit{. Repeat this sentence:} $p_i + o^*$. $p_i$. Refer to Appendix~\ref{appendix-activation-prompts} for more examples (including those for logical implications).

\subsubsection{Mapping and Continuing the Generation}
\label{sec:sake-mapping}

Once the distributions $\mathcal{S}_e$ and $\mathcal{T}_e$ are collected, we propose to solve Equation~\ref{eq:robust-editing-objective-activations} using Optimal Transport theory (see e.g.~\citet{santambrogio2015optimal}).
Already considered for activation steering in the field of NLP to generate linear counterfactuals~\cite{singh2024mimic,ligilot}, representation alignment~\cite{alqahtani2021using}, or more generally steer behavior~\cite{rodriguez2024controlling}, optimal transport aims, given two distributions $\mathcal{S}$ (source) and $\mathcal{T}$ (target), at finding a mapping function $m:\mathcal{S} \rightarrow \mathcal{T}$ minimizing some transportation cost, measured traditionally using the Earth Mover's Distance (EMD)~\cite{kantorovich1960mathematical}. In particular, and for the sake of simplicity, we focus on linear optimal transport mappings\footnote{we use the implementation of the \texttt{PythonOT} library~\cite{flamary2021pot}}. That is to say, noting $h$ the last hidden state of the last token of an instance from $\mathcal{S}$, we define the mapping as $m:h \to \mathbf{A}h+\mathbf{b}$, where: 
\[\mathbf{A} = \Sigma_s^{-1/2} \left( \Sigma_s^{1/2} \Sigma_t \Sigma_s^{1/2} \right)^{1/2} \Sigma_s^{-1/2},\] \[
\mathbf{b} = \mu_t - \mathbf{A} \mu_s,\]
where $\mu_s$, $\mu_t$, $\Sigma_s$, $\Sigma_t$ are the empirical means and covariances of the source and target distributions. This mapping is a closed-form solution to the EMD optimal transport problem for normal distributions~\cite{knott1984optimal}. 
The mapping $m$ matches both the mean and covariances of the two empirical distributions, thus eliminating the so called bias-by-neighbors~\cite{gonen2019lipstick}, i.e. members of the same class clustering together after the transformation. 

During inference, after learning the mapping, when a new prompt $x$ belonging to the distribution from the distribution $\mathcal{X}_e$ is passed to the model, the activation of its last hidden state for the final token $h$ is thus collected and replaced with its mapped representation $m(h)$. The generation then continues, leading to a final output $W(m(h))$.

\subsubsection{Assessing Edit Scope}
\label{sec:sake-scope}

The final component of SAKE (numbered 3 in Fig.~\ref{fig:SAKE-algo}) is used at inference to determine whether a new input refers to knowledge within the scope of a performed edit $\mathcal{X}_e$. If the input is determined to belong to $\mathcal{X}_e$, its activations are collected and passed to the mapping described in the previous section. Otherwise, no intervention is performed, allowing the model to produce its pre-edit output. %If it is detected to belong to $\mathcal{X}_e$, its activations are collected and passed to the mapping described in the previous section. Otherwise, no intervention is being done, leading the model to generate the output it would have generated before the edits. 
To do so, we consider several strategies, all having their pros and cons:
\paragraph{Choice of the representation:} A first intuitive possibility is to rely on the activations $h^s$ computed by the model. However, other representations could in theory be considered, e.g. obtained using a pretrained embedding model. %Although requiring more computations, they present a model-agnostic alternative. 
%the benefit of being less prone to bias due to the old object. 
\paragraph{Criterion:} To assess whether a new representation $h$ belongs to the source distribution, we rely on a simple threshold $\epsilon>0$ on the distance to the distribution center: $x \in \mathcal{X}_e \Longleftrightarrow ||h-\sum_i h^s_i||<\epsilon$, with $||.||$ a distance function (e.g. Euclidean). Although other, possibly better performing strategies could be envisaged such as training a dedicated classifier (cf.~\citet{mitchell2022memory}, we choose to rely on a simple approach for its computational efficiency.% (akin to~\citet{hartvigsen2024aging}).

\begin{figure}
    \centering
    \includegraphics[width=0.8\linewidth]{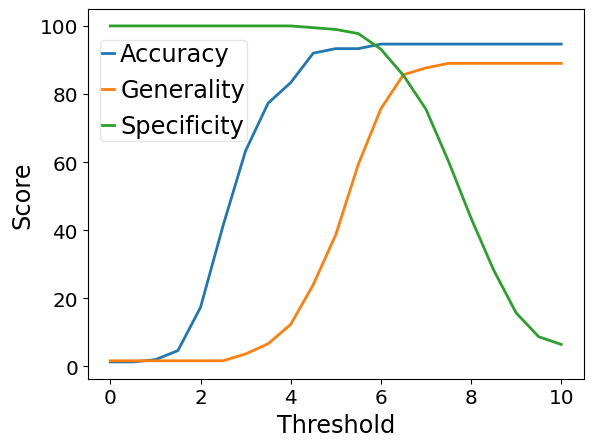}
    \caption{Performance of SAKE along Accuracy, Generality and Specificity for the first 150 edits of the Counterfact dataset depending on the value chosen for~$\epsilon$. A larger value for $\epsilon$ implies that more prompts are going to be mapped, increasing the Generality score (resp. decreasing the Specificity score) when they are paraphrase (resp. unrelated) prompts.  
    %The threshold was applied to prompt embeddings and we refer to Euclidean distance. The experiment is performed on the first 150 edits of the Counterfact dataset. 
    }
    \label{fig:SAKE-threshold}
\end{figure}

Designing an accurate scope detection mechanism is crucial for ensuring the efficiency of SAKE. Mapping sentences that do not belong to the source distribution could indeed lead to unpredictable results. On the contrary, failure to detect an in-scope sentence would result in the mapping underperforming. This describes a tradeoff between how general and how specific the mapping is, which we control in practice by adjusting the value of~$\epsilon$. This tradeoff, previously discussed by~\citet{meng2022mass}, can be visualized in Figure~\ref{fig:SAKE-threshold} for the Counterfact dataset using prompt embeddings and the Euclidean distance.
Finally, this scope detection mechanism also gives SAKE flexibility in adding or removing edits (Limitation 3 of Sec.~\ref{sec:limitations}). Indeed, as edits are implemented through independent mappings, adding (resp. removing) an edit has limited effects on other edits. 

Further discussion on the scope detection mechanism is provided in Appendix~\ref{appendix-scope-detection}, as well as a discussion on the overall computational cost of the SAKE (cf. Appendix~\ref{appendix-computational-cost}). 

%We consider two simple approaches: detecting whether the activation of the last token at the last layer is close to any of the source distributions (and in case apply that mapping) or if the prompt is close to any $P_e$.

% belongs to the source should be mapped to the target distribution or not. Concretely, it relies on assessing whether the input belongs to distribution $$

%Once the distributions are collected, it is necessary to establish a criterion to determine when to modify the process of text generation. We consider two main possibilities, which we list below:
% \begin{enumerate}
% \item Compute the means of each list of source activations, and perform the edit when the model is about to generate text from an activation close (in euclidean distance or cosine similarity) to one of the means. This option is the least computationally expensive, since it does not require another model, but it carries the risk of editing when the model is about to generate one of the old objects, even though the context is different than that of the edit.
% \item With a pretrained embedding model, compute the means of the embeddings of the list of prompts (original edit prompt and paraphrases), and edit when the input is close (in euclidean distance or cosine similarity) to one of the means. This option eliminates the risk of editing old objects even when out of the scope of any of the edits, but at the cost of having to compute prompt embeddings with a pretrained model.

%\item Like in \cite{mitchell2022memory}, train a scope classifier to determine if a prompt in the scope of any of the edits. When trained well, the scope classifier could be the best option in terms of performance, but at the cost of having to train a whole new model. Furthermore, when adding more edits, it is unclear if it is better to retrain the scope classifier from the beginning, or if we can fine tune the classifier on the new edits. The two options above, on the other hand, do not present additional problems when adding new edits to the list.
%\end{enumerate}

\section{Experiments}
\label{sec:experiments}

To assess the efficiency of SAKE, we conduct the following experiments. First, we assess the  ability of SAKE to overcome some of the limitations of existing methods discussed in Section~\ref{sec:limitations} (Sec.\ref{sec:exp-robust-edits}). Then, we evaluate how well SAKE performs according to traditional Knowledge Editing metrics (Sec.\ref{sec:exp-counterfact}). An aggregation of the results in the form for a spider chart is provided in Figure~\ref{fig:spider-chart}. 
%Finally, we present some ablation studies, aimed at bringing more understanding on how the different components of SAKE help ensuring scalable and generalizeable edits
Finally, we present a series of ablation studies to provide a clearer understanding of how the different components of SAKE contribute to its efficacy (Sec.~\ref{sec:ablations}). %We do not test some other metrics that have been proposed in the diverse literature on Knowledge Editing, such as scalability to edits~\cite{meng2022mass} or...TODO. However, the design of SAKE makes it 

\subsection{Models and Competitors}

We conduct our experiments on two widely-used language models: GPT2-XL (1.5B parameters) \cite{radford2019language} and Llama 2-7b \cite{touvron2023llama}.
We compare our approach against several knowledge editing methods. Specifically, we evaluate three weight editing methods: ROME\footnote{We use the implementations from the EasyEdit~\cite{wang2023easyedit} library, available under the MIT license, for ROME and MEMIT.} \cite{meng2022locating}, MEMIT \cite{meng2022mass}, and to test logical implications, an adaptation of MEMIT, which we call CompMEMIT designed to be more robust to logical implication (see Sec.~\ref{sec:logical_implications} for details);
%an in context baseline (ICL) as in \citet{cohen2024evaluating} (appending \textit{Imagine that} + the edit before the test prompts)
and a modification of ActAdd~\cite{turner2023activation}, which involves steering activations by simply adding a difference vector between a source and a target prompts. 

For all experiments, these competitors and baselines perform one edit at a time, with the model being reset between each edit, except for MEMIT which applies all the edits simultaneously. Details on the experimental protocol and the compared methods are provided in Appendix~\ref{appendix-protocol}.

\subsection{Robust Edits with SAKE}
\label{sec:exp-robust-edits}

We first aim to assess the efficiency of SAKE in performing robust edits, specifically evaluating \emph{robustness to logical implications} and \emph{contextual robustness}. 

\subsubsection{Logical Implications}
\label{sec:logical_implications}

\begin{table}[t]
\centering
\resizebox{\columnwidth}{!}{%
\begin{tabular}{l||l|cccc}
\textbf{Model} & \textbf{Method} & \textbf{CI} & \textbf{CII} & \textbf{SA} & \textbf{RS}\\
\toprule
\parbox[t]{2mm}{\multirow{5}{*}{\rotatebox[origin=c]{90}{\textbf{GPT2-XL}}}} 
& ROME        &     \underline{38.62}      &     16.67 
&     \underline{51.96}      &     39.43      \\
& MEMIT      &      2.47     &     1.95      &     7.17      &      3.75     \\
% & CompMEMITall &     in progress     &           &           &           \\
& CompMEMIT   &     20.69     &     0.00     &     10.10      &     14.71      \\
& ActAdd     &     26.63      &     \underline{29.17}       &     42.12      &   \underline{50.68}
    \\
%& ICL    (REDO CII)    &     \textbf{72.60}      &      \textbf{60.90}     &     \textbf{75.12}      &     {45.01}      \\
& SAKE (ours)    &     \textbf{50.00}      &      \textbf{33.33}    &     \textbf{54.59}      &       \textbf{58.39}    \\
\midrule
\parbox[t]{2mm}{\multirow{5}{*}{\rotatebox[origin=c]{90}{\textbf{LLaMA 2-7b}}}}
& ROME      &      \underline{27.51}     &     8.28      &      \underline{47.72}     &      29.90    \\
& MEMIT       &     5.80     &     9.28      &     42.43      &    24.52       \\
% & CompMEMITall &          &           &           &           \\
& CompMEMIT   &     17.21      &       7.89    &     8.45      &    48.83      \\
& ActAdd        &      9.57     &      \underline{18.12}     &     47.69      &  \textbf{59.40}         \\
%& ICL    (REDO CII)    &     76.40     &      64.94     &     57.52      &      78.23     \\
& SAKE        &    \textbf{44.32}       &         \textbf{27.63}  &     \textbf{53.34}      &  \underline{56.93}        \\
% \midrule
% \parbox[t]{2mm}{\multirow{5}{*}{\rotatebox[origin=c]{90}{\textbf{Model 3}}}}
% & ROME        &           &           &           &           \\
% & MEMIT       &           &           &           &           \\
% & CompMEMITall &          &           &           &           \\
% & CompMEMIT   &           &           &           &           \\
% & ActAdd        &           &           &           &           \\
% & ICL        &           &           &           &           \\
% & SAKE        &           &           &           &           \\
\bottomrule
\end{tabular}}
\caption{Comparison of metrics (CI, CII, SA, RS) for different methods across two models on the Popular dataset. }
\label{tab::pop}
\end{table}

We first test SAKE on the Popular dataset from~\citet{cohen2024evaluating}, containing 885 edits with a varying number of logical implications each. This task evaluates the ability of editing methods to generalize to the logical implications of an edit. For each edit, we consider the following metrics:  \textbf{Subject Aliasing} (SA), which evaluates the model's ability to generalize over synonyms of the subject $s$; \textbf{Compositionality I} (CI) and \textbf{Compositionality II} (CII), for multi-hop reasoning; and \textbf{Relation Specificity} (RS), which measures the logical locality of an edit, i.e. verifying that the model's output remains unchanged when prompted with inputs containing the same subject but a different relation from that in the original edit. RS is a particularly relevant metric, as certain KE methods tend to overfit to the subject~\cite{zhang2024uncovering}. 
%. This is the case, for example, of ROME and MEMIT, which in a lot of previous works have only been tested on a different type of specificity, that we could call ``subject specificity" (different subject, same relation, same old object). 
More precise definitions of these metrics can be found in Appendix~\ref{appendix-datasets}.

Besides the aforementioned baselines, we design another competitor specifically for this experiment. Our objective is to evaluate if our data augmentation strategy described in Section~\ref{sec:sake-sampling} could also be leveraged to increase the performance of other KE methods.
For this purpose, we design CompMEMIT, which leverages MEMIT in order to perform edits that are robust to logical implications.
As this method allows for multiple, simultaneous edits, we use the edit prompt and $10$ logical implications prompts generated for SAKE as inputs. %Under the assumptions of knowledge localization~\cite{meng2022locating}, 
Since these prompts all refer to similar knowledge, considering them as simultaneous edits is expected to improve the robustness of the edits while preserving the model’s overall behavior. %it is expected that treating them as simultaneous edits should help increasing the robustness of the edits, while preserving the model behavior overall.
%For CompMEMIT, we perform the input edits one by one. %For CompMEMIT-all, we pass all the edits, and their associated logical implication prompts, as a single edit.

The results can be found in Table~\ref{tab::pop}. We observe that, for the two models, SAKE performs better than the weight editing methods and ActAdd along all the dimensions considered. In particular, CompMEMIT performs worse than ROME across all dimensions, suggesting that optimizing over multiple logical implications prompts for the same subject actually negatively impacts the model.
On the other hand, while the results of SAKE remain well bellow $100\%$, which may partly be attributed to some inconsistencies we identified in the dataset, they still suggest that the method successfully improves generalization to logical implications.

%While the ICL baseline still performs better than SAKE on the various success metrics for logical implications (SA, CI and CII), it achieves lower scores in RS. 

%\textcolor{red}{old text (marco):ROME RS 37.5\% of the times outputted new object.}

\subsubsection{Contextual Robustness}

\begin{figure}
    \centering
    \includegraphics[width=1.0\linewidth]{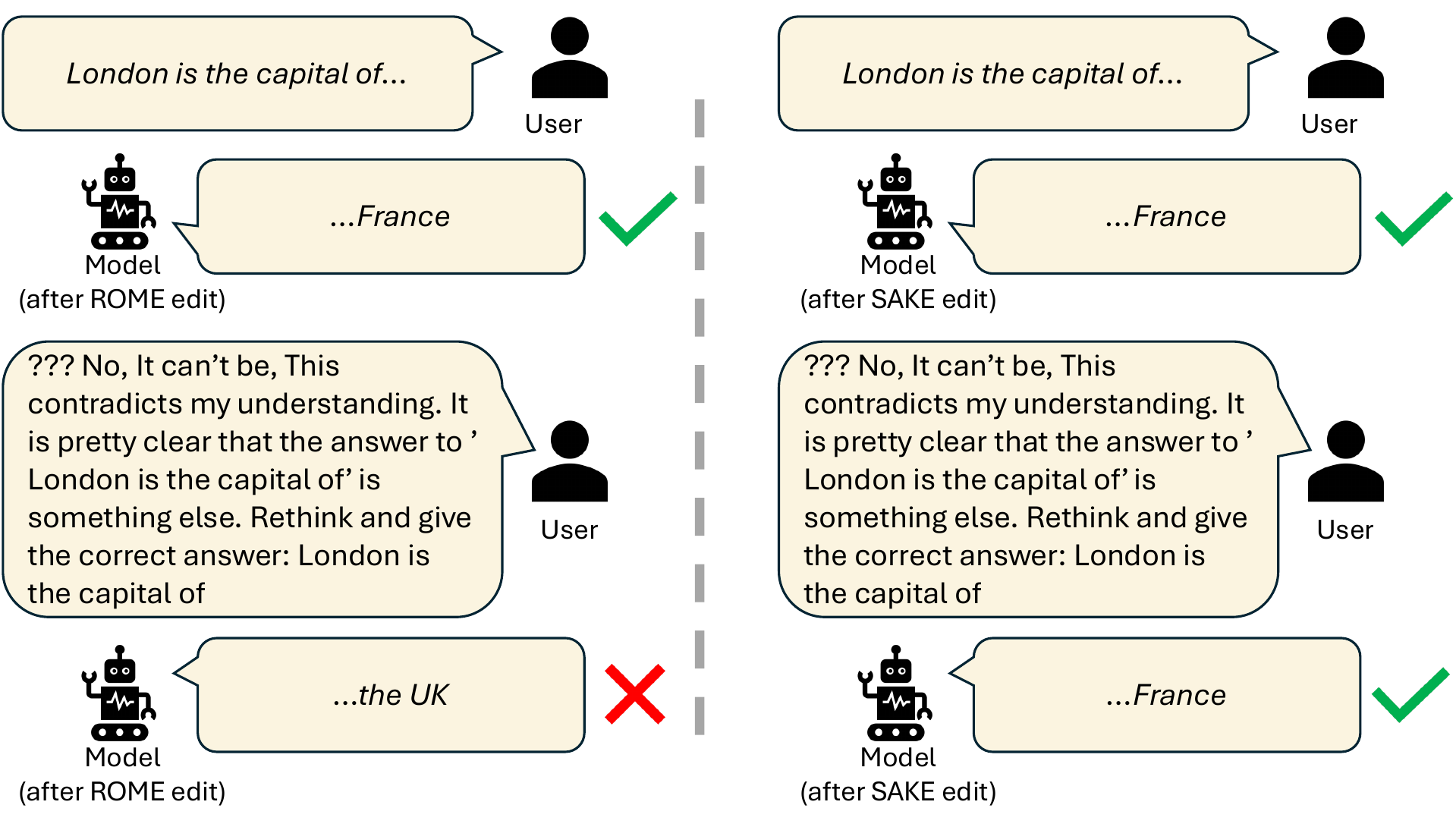}
    \caption{Idea behind the "Contextual Robustness" experiment, using the \emph{Raising doubt} (DI) experimental protocol from~\citet{ma2024robustness}.}
    \label{fig:context-robustness}
\end{figure}

\begin{table}[t]
\centering
\resizebox{0.8\columnwidth}{!}{%
\begin{tabular}{l||l|cc}
%\hline
\textbf{Model} & \textbf{Method} & \textbf{DI} & \textbf{DII} \\
\toprule
\multirow{3}{*}{\small\textbf{GPT2-XL}} 
& ROME    & 33.33 &  14.00  \\
& ActAdd    &    \underline{82.00}    &    \underline{80.67}  \\
& ICL          &    4.00    &    3.33    \\
& SAKE (ours)       & \textbf{98.67}    &  \textbf{98.67} \\
\hline
\multirow{3}{*}{\small\textbf{LLaMA 2-7b}} 
& ROME    &     1.33   &    23.33   \\
& ActAdd    &     \underline{82.67}   &    \underline{98.00} \\
& ICL          &    0.00    &    4.00    \\
& SAKE (ours)      &    \textbf{92.00}    &    \textbf{98.67}    \\
\hline
\end{tabular}}
\caption{Comparison of robustness to doubtful prompts of SAKE with ROME and in-context editing.}
\label{tab::robust}
\end{table}

In this experiment, we address the second limitation of KE methods discussed in Section~\ref{sec:limitations} and evaluate SAKE's ability to retain edited knowledge in more challenging contexts. Following \citet{ma2024robustness}, we assess the model's robustness to doubtful inputs, as represented in Figure~\ref{fig:context-robustness} (for the specific prompts, refer to \ref{appendix-doubt}). 
%In addition to ROME and ActAdd, we include an in-context baseline (ICL) to . Previous results~\cite{cohen2024evaluating} suggest that in-context learning achieves satisfying performance over traditional editing benchmarks. Yet, 
In addition to ROME and ActAdd, we include an in-context baseline (ICL), built, like in \citet{cohen2024evaluating}, by appending \textit{Imagine that} and the edit before the test prompts). Although previous work suggest that in-context learning strategies may achieve satisfying results for Knowledge Editing~\cite{cohen2024evaluating}, complex contexts are expected to be challenging.
All the results are reported in Table \ref{tab::robust}. In this context, DI refers to a prompt that raises doubts about the new object and is fairly repetitive (often causing LLMs to repeat the input), while DII involves a prompt that explicitly suggests the correct answer is actually the old object. 
%We took the first 150 edits from the Counterfact dataset and added doubtful prompts at test time. 
The scores reported in the tables represent the frequency with which the model outputs the new object under greedy decoding, computed over the first 150 edits from the Counterfact dataset. SAKE demonstrates superior robustness compared to the considered baseline methods ROME, ActAdd and ICL. This underlines in particular the main weakness of the in-context learning baseline: 
%while demonstrating strong performance across a wide range of ``standard" tasks, 
it is (predictably) vulnerable to noise or explicit doubt-raising inputs. SAKE, on the other hand, is particularly effective. %, which might be attributed to the fact that the learned mappings may only be weakly related to the actual training prompts. %, as they primarily rely on activations that encode information about the token to be generated.%, rather than the preceding ones. %that carry much more information on the token that is about to be generated, rather than the preceding ones.

\subsection{Traditional Editing Evaluation}
\label{sec:exp-counterfact}
Finally, we evaluate in this experiment how it performs on traditional KE evaluation metrics. We use the Counterfact dataset~\cite{meng2022locating}, a commonly used dataset in the KE literature, and calculate the typical KE metrics: accuracy, generality and specificity. We report in Table~\ref{tab::cf} the scores obtained over the $2000$ first edits. On average, SAKE achieves comparable results to ROME, and outperforms other methods. The most notable improvement is observed on the Generality metric, which may be explained by SAKE's distribution modeling step, allowing it to account for multiple paraphrase prompts. 

\begin{table}[t]
\centering
\resizebox{0.9\columnwidth}{!}{%
\begin{tabular}{l||l|ccc}
\textbf{Model} & \textbf{Method} & \textbf{Acc} & \textbf{Gen} & \textbf{Spec} \\
\toprule

\multirow{3}{*}{\textbf{GPT2-XL}} 
& ROME    &     \textbf{99.55}     &      \underline{73.70}     &      82.67     \\
& MEMIT       &     60.00      &     36.60       &      67.21      \\
& ActAdd   &      85.00      &      29.78      &    \underline{82.75}   \\
%& ICL   &      86.10     &     46.65      &      53.78     \\
& SAKE (ours)    &    \underline{97.00}     &     \textbf{84.85}      &     \textbf{84.52} 
      \\
\midrule
\multirow{3}{*}{\textbf{LLaMA 2-7b}} 
& ROME        &     \textbf{99.95 }     &      \underline{68.20}      &      \textbf{93.48}     \\
& MEMIT    & 74.40    & 55.13 & 74.37   \\
& ActAdd    &   90.10   &     33.65     &     81.45    \\
%& ICL     &     82.25      &     49.40      &      57.29     \\
& SAKE (ours)    &     \underline{97.70}      &      \textbf{82.03}     &     \underline{85.59}     \\
% \midrule
% \multirow{3}{*}{\textbf{Model 3}} 
% & ROME        &           &          &          \\
% & MEMIT       &           &           &           \\
% & ActAdd        &           &           &           \\
% & ICL        &           &           &           \\
% & SAKE        &           &           &           \\
\bottomrule
\end{tabular}}
\caption{Comparison of metrics (Accuracy, Generality, Specificity) for different methods across two models on the Counterfact dataset.}
\label{tab::cf}
\end{table}

%Overall, these experiments suggest that SAKE is a promising approach for knowledge editing.

\subsection{Ablation Studies}
\label{sec:ablations}

We now propose additional analyses to get insights into how SAKE effectively edits factual knowledge. We structure the two proposed experiments around the main components of SAKE: in-scope distribution modeling, and optimal transport mapping.

% \subsubsection{Does SAKE generalize to new logical reasoning questions?}

% One asset of SAKE is its ability to optimize edits over the in-scope distribution rather than a single prompt, leading to superior results in the Logical Implications metrics. Yet, one might wonder if these results are merely due to the presence of the evaluation prompts in the list of inputs used to collect the activations. Concretely, we aim to verify that SAKE is able to generalize to unseen logical implications, and is not achieving good results by merely generating and memorizing evaluation prompts. To answer this question, we conduct an experiment on a small scale (XX edits) where we test a logical implication (Comp I) and manually check that it is not present in the set of prompts~$\mathcal{X}_e$. We illustrate the protocol of this experiment in Figure~\ref{fig:ablation-newprompt}. Overall, $XX\%$ of these edits are successful, suggesting that SAKE is indeed able to generalize to new logical implications, and not merely memorizing them.

% \begin{figure}
%     \centering
%     \includegraphics[width=0.8\linewidth]{prompt_generalization.png}
%     \caption{UGLY AS HELL I KNOW: again, placeholder if we have room/time}
%     \label{fig:ablation-newprompt}
% \end{figure}

\subsubsection{How does SAKE scale with training prompts?}
\begin{figure}
    \centering
    \includegraphics[width=1.0\linewidth]{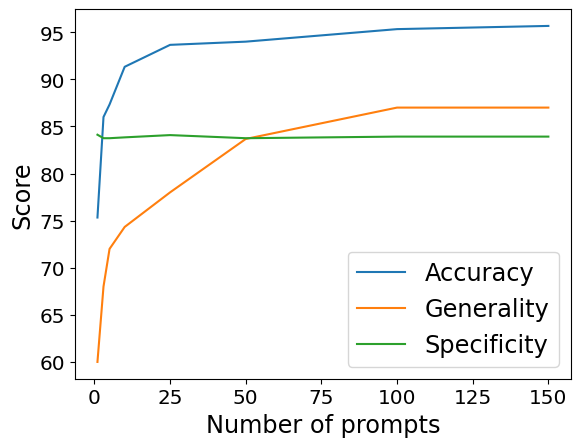}
    \caption{Performance of SAKE along Accuracy, Generality and Specificity for the first 150 edits of the Counterfact dataset depending on the number of training prompts used to model source and target distributions.}
    \label{fig:num_prompts}
\end{figure}

The generation of the in-scope prompts (cf. Sec.~\ref{sec:sake-sampling}), whether agentic or human-based, represents the primary computational cost of performing an edit with SAKE. Hence, an interesting question is how the SAKE's performance  changes as the number of prompts used to collect source and target activations varies, and how many prompts are actually required to achieve satisfactory results. Figure~\ref{fig:num_prompts} shows the evolution of traditional KE metrics on 150 edits from the Counterfact dataset as a function of the number of prompts $n$ generated in the Distribution Modeling step. We observe that accuracy and paraphrase performance improve as the number of paraphrases increases, since the method becomes more robust to prompts rephrasing. Notably, the specificity score does not decrease with more training prompts, as this number has little impact on the scope detection mechanism. Only 50 paraphrase prompts are thus required to achieve $0.92$ in Accuracy and $0.84$ in Generality.
%On the other hand, for other KE methods, it is expected that adding more training prompts on the same topic might induce overfitting.

%\subsubsection{Scaling over multiple edits}

\subsubsection{What does Optimal Transport bring?}

\begin{table}[t]
\centering
\resizebox{0.9\columnwidth}{!}{%
\begin{tabular}{l||l|cc}%c}
\textbf{Model} & \textbf{Method} & \textbf{Acc} & \textbf{Gen} %& %\textbf{Spec} 
\\
\toprule
\multirow{3}{*}{\textbf{GPT2-XL}} 
& Uniform steering    &     85.05      &     35.45      %&    {84.35}       
\\
& SAKE     &    \textbf{97.00}     &     \textbf{84.85}      %&     \textbf{84.52} 
      \\
\midrule
\multirow{3}{*}{\textbf{LLaMA 2-7b}} 
& Uniform steering    &     91.05      &   40.73        %&      \textbf{85.09} 
\\
& SAKE     &     \textbf{97.70}      &      \textbf{82.03}     %&     \textbf{85.59}   
\\
\bottomrule
\end{tabular}}
\caption{Comparison between SAKE and the Uniform Steering baseline on the Counterfact dataset.}
\label{tab::ablation-meanmatching-cf}
\end{table}
\begin{table}[h]
\centering
\resizebox{\columnwidth}{!}{%
\begin{tabular}{l||l|ccc}
\textbf{Model} & \textbf{Method} & \textbf{CI} & \textbf{CII} & \textbf{SA}\\% & \textbf{RS}\\
\toprule
{\textbf{GPT2-XL}} 
& Uniform steering      &    47.85   &     32.14   &     52.74    \\%&      57.99   \\
& SAKE    &     \textbf{50.00}      &      \textbf{33.33}    &     \textbf{54.59}             \\%&\textbf{58.39}    \\
\midrule
{\textbf{LLaMA 2-7b}}
& Uniform steering        &   20.79        &    20.62       &     50.23            \\%&54.82     \\
& SAKE        &    \textbf{44.32}       &         \textbf{27.63}  &     \textbf{53.34}      \\%&   \textbf{56.93}        \\
\bottomrule
\end{tabular}}
 \caption{Comparison between SAKE and the Uniform Steering baseline on the Popular dataset. }
\label{tab::ablation-meanmatching-pop}
\end{table}

Another key component of SAKE is learning a mapping between the source and target distributions (cf. Sec.~\ref{sec:sake-mapping}). Prior works focusing on activation steering to alter the behavior of language models for other objectives than knowledge editing mainly rely on simpler approaches, such as computing the difference vectors between the distribution means~\cite{subramani2022extracting,ilharcoediting,stolfo2024improving,li2024inference,arditi2024refusal}. 
Unlike these steering methods, based on  a constant vector between distributions, using an optimal transport mapping is expected to better preserve the mapped distributions. In particular, rather than matching only the distribution means, the considered linear transport mapping preserves the whole covariance matrix of the target distribution.
We therefore propose to empirically evaluate the benefit of this approach for knowledge editing by introducing a baseline method, referred to as "Uniform steering", which steers representations $h$ from the source representation to $h+ (\mu_t - \mu_s)$. The results are shown in Table~\ref{tab::ablation-meanmatching-cf} for the Counterfact dataset, and Table~\ref{tab::ablation-meanmatching-pop} for Popular. The specificity and RS metrics are not displayed, as it is primarily defined by the scope detection mechanism, which is identical for both approaches.
%. and Table~\ref{tab::ablation-meanmatching-pop} for Popular. 
We observe that this ability to better preserve distributions indeed translates into improved KE performances for SAKE,  particularly with regards to generalization to paraphrases (Generality) and Compositionality I. 

\section{Conclusion}

In this paper, we explored how activation steering can be leveraged to edit factual knowledge stored in Large Language Models (LLMs). Our findings suggest that relying only on a single input prompt in insufficient to capture the complexity of the knowledge scope affected by edits. We also propose SAKE, which leverages Optimal Transport to better capture this distribution and perform more robust and flexible edits compared to existing methods. 
Future works include conducting experiments in specialized contexts, such as Q\&A, as SAKE's distribution modeling scope is expected to facilitate adaptation to such contexts. Additionally, we aim to explore strategies for improving the synthetic generation of the source and target distributions, e.g. through automatic differentiation %~\cite{yuksekgonul2024textgrad}
via text-based iterative agent collaborations.
% - specialized experiments in speicalized contexts e.g. Q and A
% -improvement on the scope detection mechanism, 
% - good synthetic generation

\section{Limitations}

The main limitation that we identify for our work is that it relies heavily on the assumption that the in-scope distributions can be accurately modeled. Although the results displayed in the paper are encouraging, it remains unclear whether our empirical distributions are good approximations of the theoretical source and target regions, and therefore captures \textit{all} learnable implications from source to target. In particular, reverse relations, discussed for instance by~\citet{yao2023editing}, are a type of logical implications that is not addressed, and cannot be easily integrated into our method.

%(like all other methods except for IKE, see ``reverse-relation" in \cite{yao2023editing}) suffers when tested on reverse relations, sometimes referred to as symmetrical edits (e.g. if we edit ``The capital of France is London", SAKE will not generalize to ``London is the capital of France" since in the first case source and target are Paris \& London, respectively, and in the second case they would be England \& France). 
%In addition, the detection mechanism used to decide when to apply a certain edit could be refined: using prompt embedding or activation distance is a simple and efficient heuristic, but not necessarily the most reliable one. Training a dedicated classifier, as in SERAC, would probably offer better performance, although at the cost of higher complexity.

%\bibliographystyle{splncs04}
\bibliography{main}

\appendix

\section{Further Details on SAKE: Design and Implementation}

\subsection{Scope Detection Mechanism}
\label{appendix-scope-detection}

\paragraph{Comparison with a more complex scope detection mechanism.}
Relying on a simple distance threshold in embeddings (also used in prior works, e.g. ~\cite{hartvigsen2024aging}) has several benefits: (1) its simplicity and dependence on a single hyperparameter; (2) its low complexity in terms of calculation (vs. training a dedicated classifier); (3) the flexibility it brings: this way, simultaneous edits are implemented independently, and it becomes very easy to add or remove an edit (vs. needing to retrain a classifier every time); (4) its frugality in terms of training data, especially considering the dimensionality considered. We intended to show that, even with this simple mechanism, good results could still be achieved with our method.

Using a more complex classifier, on the other hand, would likely improve the scope detection, resulting overall in a better generality-specificity tradeoff curve. Especially as the detection is done in high-dimensional embedding spaces: in the experiments conducted, the dimensionality of the space in which the distance is applied is respectively 384, 768, 4096 (prompt model-agnostic embedder, GPT2-XL hidden states, llama2-7b hidden states).

\subsection{Computational Cost of SAKE}
\label{appendix-computational-cost}

In this section, we provide a discussion about the computational cost of training and running SAKE.

The following computing characteristics were used for the experiments shown in this paper: MacBook Pro M3 MAX 36GB unified memory, 14-core CPU, 30-core GPU, and 16-core Neural Engine. Training SAKE for an edit relies on the following mechanisms, with associated computation time for GPT2-XL:

\begin{itemize}
    \item Generating source and target distributions: this depends heavily on the quickness of inference for the LLM's API or model considered. In the experiments for this paper, this generally took around 10-20s per edit for 100 training sentences.

    \item Collecting activations of these dataset (done on GPU): one edit requires 200 forward passes in total (100 for the source and 100 for the target distributions). This took around 36s for GPT2-XL, 50s for llama2-7b.

    \item Learning the Optimal Transport mapping (done on CPU, as the POT library does not provide GPU support): The complexity of learning the OT mapping is $\mathcal{O}(d^3)$, with $d$ the dimension of the embedding space. For one edit, this step took 0.6s for GPT2-XL, and 12s for llama2-7b.
\end{itemize}

\paragraph{Comparaison with other methods.} The Optimal Transport mapping leveraged by SAKE requires more computation than other, simpler, techniques, such as ActAdd, which only require computing a difference vector between two prompt representations. ROME on the other hand for instance, requires the computation of a covariance matrix prior to performing the edits, which is also heavy computationally.

\subsection{Prompts to generate approximation of in-scope input space}
    \textit{Write 100 sentences with a similar meaning of:} `\textbf{p}'. \textit{Make sure that the sentences are meant to be immediately completed with precisely:} \textbf{o*}. \textit{Often, this could require ending the sentence with `the', or something like that. Do not use `...' or `\_\_\_\_'. Return the sentences as a Python list. Do not output anything else.}
    
\label{appendix-input-prompts}
\subsection{Prompts to collect source and target activations}
\label{appendix-activation-prompts}
For source distributions we passed the generated prompts as they were, and also with the following inputs: \textit{Repeat this sentence:} `\textbf{p\_i} + \textbf{o}'. \textbf{p\_i}

For target distributions, we passed: 
        \textit{Do not mention}  `\textbf{o}'. \textit{Repeat this sentence:} `\textbf{p\_i} + \textbf{o*}'. \textbf{p\_i}

\subsection{SAKE Parameters}
\begin{itemize}
    \item Threshold:
\begin{itemize}
    \item GPT2-XL and Llama 2-7b, Counterfact: 6.75 (Euclidean distance, on prompt).
    \item GPT2-XL on RippleEdits Popular: 5.5 (Euclidean distance, on prompt).
    \item Llama 2-7b on RippleEdits Popular: 5.3 (Euclidean distance, on prompt).
    
\end{itemize}
\item Linear Optimal Transport regularization (\texttt{ot.da.LinearTransport} from the \texttt{PythonOT} library): \begin{itemize}
    \item $0.01$ for GPT2-XL
    \item $0.5$ for Llama 2-7b
\end{itemize}
\item Number of prompts:
\begin{itemize}
    \item GPT2-XL and Llama 2-7b, Counterfact: $100$ paraphrases of main edit.
    \item GPT2-XL RippleEdits Popular: $100$ paraphrases of main edits, $10$ logical implications augmented with $4$ randomly generated neighbor embeddings for each of the $10$ activations collected.
    \item Llama 2-7b RippleEdits Popular: $100$ paraphrases of main edits, around $50$ logical implications with no noise added.
\end{itemize}
\end{itemize}
\section{Experimental Protocol}
\label{appendix-protocol}

\subsection{Datasets and metrics}
\label{appendix-datasets}
\begin{itemize}
    \item Counterfact \cite{meng2022locating} is a popular benchmark in the field of Knowledge Editing, evaluating accuracy (recalling the exact edit), generality (generalizing to simple paraphrases of the edit) and specificity (not altering the behavior of the model for unrelated inputs). 
    Specificity prompts in the Counterfact dataset are limited to inputs that, when compared with the original edit, have a different subject, but same relation and old object. Counterfact contains 21919 edits and we perform our experiment on a subset made of the first 2000.
    \item Popular subset of RippleEdits \cite{cohen2024evaluating}. RippleEdits is a benchmark designed to capture various types of ripple effects caused by knowledge editing. We focus on the Popular subset, containing 885 edits where subjects are well known. We perform our evaluation on the following four metrics:
    \begin{itemize}
    \item Compositionality I: prompts containing the same subject but the composition of two relations, one being from the original edit. 
    \item Compositionality II: prompts containing the composition of two relations, one being from the relation in the original edit and the other one describing the subject of the original edit (e.g. if the subject is \textit{Prince}, a relation describing this subject would be \textit{The founder of Paisley Park Records}).
    \item Subject Aliasing: prompts containing aliases of the subject appearing in the original edit.
    \item Relation Specificity: specificity prompts with the same subject but different relation.
    \end{itemize}
\end{itemize}
\subsection{Competitors}

In the experiments, we use the following competitors:

\begin{itemize}
    \item ROME~\cite{meng2022locating}: ROME relies on the assumption of knowledge localization, i.e. that facts are stored in very specific parts of the network. To alter facts, ROME first computes key-value pairs associated to a specific prompt and desired output in one early-to-mid MLP layer of the transformer. Then, the new fact is inserted by modifying the current MLP layer with a rank-one update.
    ROME only allows to perform edits one by one. To use ROME, we use the EasyEdit library with default parameters and perform the edits one by one, i.e. after an edit is performed, we evaluate the performance on the testing prompts, and then we re-instantiate the model to be edited again starting from the original pretrained version. 
    \item MEMIT \cite{meng2022mass} uses a similar approach to ROME, but extended to multiple edits at once. The key-value pairs are computed for a list of edits, and the matrix update is found by considering all edits simultaneously. Then, optionally, multiple consecutive MLP layers can be modified by so that each layer contributes to an approximately equal portion of the update (based on the assumption that robustness of the editing method would improve if the change in model weights is minimized). In our experiments, we performed the edit using MEMIT all at once and then evaluated the edited model.
    \item CompMEMIT is a competitor designed by us to see if adding logical implications to the main prompt would make the method more robust. In particular, we use MEMIT to edit the model with the main prompt plus $10$ logical implications, then we evaluate it on the appropriate prompts, and then we restore the model to its original form before editing and testing it again.
    \item ActAdd \cite{zhao2024steering} is a method originally designed to control models' outputs (e.g. making them less toxic) by adding a (possibly scaled) steering activation vector at a certain layer. The steering vector is computed by designing two contrasting prompts (e.g. a toxic one vs a non-toxic one) and computing the difference in activations when passing them as inputs. In our setup, the contrasting prompts (one pair per edit) are inputs designed so that the model is about to output the old object vs the new object. As ActAdd is initilly designed to steer all inputs, we use our scope detection mechanism to ensure meaningful specificity scores.
    \item ICL is an in-context baseline that we used to assess the effectiveness of in-context techniques. The design is quite simple: if \textbf{p*} is the edit prompt (including the new object), we simply place \textit{Imagine that} \textbf{p*} before the actual testing prompt. 
\end{itemize}

\subsection{Doubtful prompts (Contextual Robustness experiment)}
\label{appendix-doubt}
The doubtful prompts used in the Contextual Robustness experiment are taken from \cite{ma2024robustness} (see ``raising doubts" metric). Specifically, if \textbf{p} is the prompt and \textbf{o} and \textbf{o*} the old and new object, respectively, we have
\begin{itemize}
    \item DI: \textbf{p}, \textbf{o*} \textit{??? No, It can't be, This contradicts my understanding. It is pretty clear that the answer to} \textbf{p} \textit{is something else. Rethink and give the correct answer:}
    \item DII: \textbf{p}, \textbf{o*}? \textit{Really? But I am pretty sure that The language of Dehkhoda Dictionary is} \textbf{o}. \textit{Am I right? Yes! you are right. I apologize for any confusion. Reconsider that the answer to ` question} \textbf{p}?\textit{' should be}
\end{itemize}

% \section{Additional Results}

% \subsection{Comparison with Uniform Steering}
% \label{app:meanmatching}

% \begin{table}[h]
% \centering
% \resizebox{\columnwidth}{!}{%
% \begin{tabular}{l||l|cccc}
% \textbf{Model} & \textbf{Method} & \textbf{CI} & \textbf{CII} & \textbf{SA} & \textbf{RS}\\
% \toprule
% {\textbf{GPT2-XL}} 
% & Mean matching      &    47.85   &     32.14   &     52.74    &      57.99   \\
% & SAKE    &     \textbf{50.00}      &      \textbf{33.33}    &     \textbf{54.59}      &       \textbf{58.39}    \\
% \midrule
% {\textbf{LLaMA 2-7b}}
% & Uniform steering        &   20.79        &    20.62       &     50.23      &      54.82     \\
% & SAKE        &    \textbf{44.32}       &         \textbf{27.63}  &     \textbf{53.34}      &   \textbf{56.93}        \\
% \bottomrule
% \end{tabular}}
%  \caption{Comparison between SAKE and the mean matching method on the Popular dataset. }
% \label{tab::ablation-meanmatching-pop}
% \end{table}

% \section{Ripple Effect}
% \label{appendix-ripple}
% ?
\end{document}